\documentclass[letterpaper,twocolumn,fleqn]{article}
\usepackage{spconf,amsmath,graphicx}
\usepackage{epsfig}
\usepackage{amssymb}
\usepackage{subfigure}
\usepackage{cite}
\usepackage{multirow}
\usepackage{epstopdf}
\usepackage{graphicx}
\usepackage{amsmath}
\usepackage{color, soul}
\usepackage[style=base]{caption}
\usepackage{lineno}
\usepackage{comment}
\usepackage{array}  
\usepackage{siunitx}
\usepackage{mathtools}
\usepackage{calrsfs}
\usepackage{eucal}
\usepackage{bm}
\newcolumntype{d}{>{\displaystyle}c}
\usepackage{url}
\usepackage{hyperref}
\hypersetup{
	colorlinks=true,       % false: boxed links; true: colored links
	linkcolor=red,          % color of internal links (change box color with linkbordercolor)
	citecolor=green,        % color of links to bibliography
	filecolor=magenta,      % color of file links
	urlcolor=cyan           % color of external links
}
\usepackage{enumitem}
\usepackage{pgfplots}
\usepackage{array}
\newcolumntype{L}[1]{>{\raggedright\let\newline\\\arraybackslash\hspace{0pt}}m{#1}}
\newcolumntype{C}[1]{>{\centering\let\newline\\\arraybackslash\hspace{0pt}}m{#1}}
\newcolumntype{R}[1]{>{\raggedleft\let\newline\\\arraybackslash\hspace{0pt}}m{#1}}
\title{DurocMien: A Deep Framework for Duroc Skeleton Extraction in Constraint Environment}
\name{
	\begin{tabular}{cc}
		Akif Quddus Khan & Salman Khan  
	\end{tabular}
}
\address{
	\begin{tabular}{c}
		Department of Computer Science (IDI), Norwegian University of Science and Technology, Norway.  
		\end{tabular}
}

\begin{document}
%\ninept
%
\maketitle

\begin{abstract}

Farm animal behavior analysis is a crucial tasks for the industrial farming. In an indoor farm setting, extracting Key joints of animal is essential for tracking the animal for longer period of time. In this paper, we proposed a deep network named DUROCMIEN that exploit transfer learning to trained the network for the Duroc, a domestic breed of pig, an end to end fashion. The backbone of the architecture is based on hourglass stacked dense-net. In order to train the network, key frames are selected from the test data using K-mean sampler. In total, 9 Keypoints are annotated that gives a brief detailed behavior analysis in the farm setting. Extensive experiments are conducted and the quantitative results show that the network has the potential of increasing the tracking performance by a substantial margin. 

\iffalse
are presented in the  
Skeletanization

The paper covers the process of skeleton extraction of Pigs using the Google Tensorflow framework. Skeleton extraction and pose estimations are important for answering questions across several scientific disciplines. Advanced deep learning algorithms offer a great advantage to scientists and researches by providing results that are better in quality, are more efficient and detailed. Using these models, researches can automatically estimate the movement of the animal’s body parts from images and videos. However, existing animal pose estimation methods have limitations in speed and efficiency. Here we introduce a new code-base that uses the Stacked Densenet model to extract pig's skeleton and predict movement. We have tested the model on two datasets differentiated by the number of pigs.
\fi

\end{abstract}

\begin{keywords}
Duroc, behavior analysis, Hourglass, Stacked dense-net, K-mean sampler, . 
\end{keywords}
%

%------------------------------------------------------------------------------------------------------------------%
\section{Introduction}
\label{sec:intro}

\iffalse
This project aims to solve the problem of behavioral analysis of farm animals through movement tracking of skeletal key points on animal bodies. In order to solve this problem, we train a Deep Neural Network which learns the distribution of Keypoints across an RGB frame. In order to train the model, we annotate the data by first converting videos into individual frames, and then annotating each frame separately by specifying the important key points on the animal's body. After sufficient training, the model returns a 9x3 matrix for each frame, where each row corresponds to one keeping, the first two columns specify the x and y coordinates of the detected point, and the third column contains the confidence score of the model. After obtaining the x and y coordinates for each frame, we visualize these key points on each frame and stitch all the individual frames into a single video.

\section{Related Work}
\label{sec:related-work}
\fi

Automatic behavior analysis of different animal species is one of the most important tasks in computer vision. Due to variety of applications in the human social world like sports player analysis \cite{khan2019disam}, anomaly detection \cite{ullah2018anomalous}, action recognition \cite{yang2016vision}, crowd counting \cite{khan2019person}, and crowd behavior \cite{ullah2016crowd}, humans have been the main focus of research. However, due to growing demands of food supplies, vision based behavior analysis tools are pervasive in the farming industry and demands for cheaper and systematic solutions are on the rise. From the algorithmic point of view, other than characterization of the problem, algorithm design for humans and the farm animals are similar. Essentially, behavior analysis is a high level computer vision task and consist of feature extraction, 3D geometry analysis, and recognition, to name a few. As far as the input data is concerned, it could be obtained thought smart sensors (Radio-frequency identification \cite{maselyne2016measuring}, gyroscope \cite{ullah2019stacked}, GPS \cite{pray2016gps}). Depending on the precision of measurements, such sensors give acceptable results but they using such sensors has many drawbacks. For example, in most cases, it is required to remove the sensor from the animal to collect the data. Such process is exhausting for the animals and laborious for the human operator. Compared to this, video based automated behaviors analysis offers a non-invasive solution. Due to cheaper hardware, it is not only convenient for the animals but also cost effective for the industry. Automatic behavior analysis and visual surveillance \cite{chen2019distributed, ullah2015crowd} has been used for the security of public places (airports, shopping malls, subways etc.) and turned into a mature field of computer vision.

\begin{figure*}[htp]
	\centering
	\includegraphics[scale=0.32]{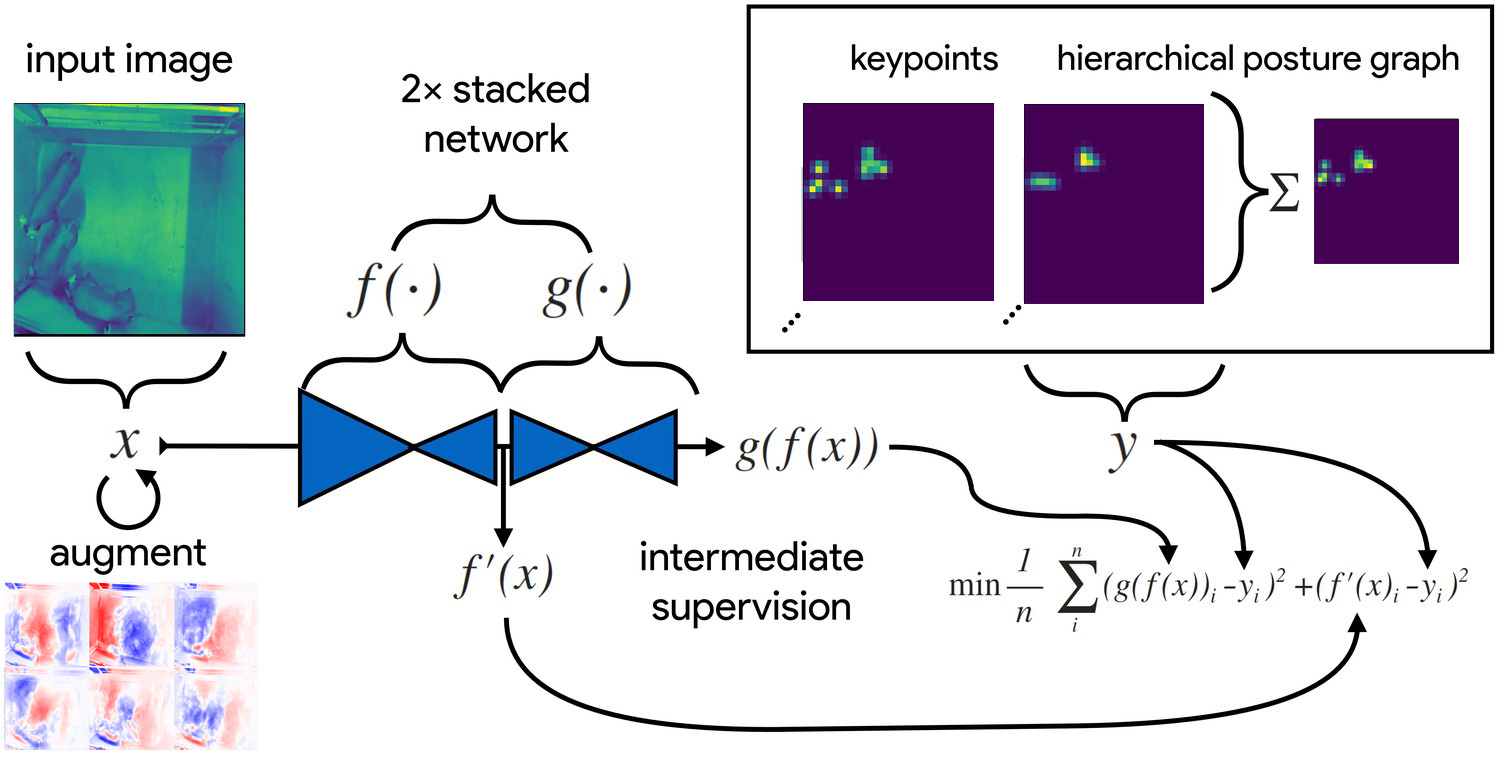}
	\caption{Detailed Illustration of Model Training Process}
	\label{fig:lossfunction}
\end{figure*}

In this regard, Hu et al. \cite{hu2017maskrnn} proposed a recurrent neural network named MASK-RNN for the instance level video segmentation. The network exploit the temporal information from the long video segment in the form of optimal flow and perform binary segmentation for each object class. Ullah et al. \cite{ullah2017human} extracted low level global and Keypoint features from video segments to train a neural network in a supervised fashion. The trained network classify different human actions like walking, jogging, running, boxing, waving and clapping. Inspired from the human social interaction, a hybrid social influence model has been proposed in \cite{ullah2018hybrid} that mainly focused on the motion segmentation of the moving entities in the scene. Lin et al. \cite{lin2017feature} proposed a features pyramid network that extract features at different level of a hierarchical pyramid and could potentially benefit several segmentation \cite{ullah2017density, hu2017maskrnn}, detection \cite{girshick2015fast, khan2019dimension}, and classification \cite{yu2018hierarchical, ullah2019single} frameworks. By addressing the problem of scale variability for object detection, Khan et al. \cite{khan2019dimension} proposed a dimension invariant convolution neural network (DCNN) that compliment the performance of RCNN \cite{girshick2015fast} but many other state-of-the-art object detector \cite{khan2019person, ren2017faster} could take advantage of it. Inspired from the success of deep features, \cite{ullah2019two} proposed a two stream deep convolutional network where the first stream focused on the spatial features while the second stream exploit the temporal feature for the video classification. The open-source deep framework named OpenPose proposed by Cao et al. \cite{cao2018openpose} focuses on the detection of Keypoints of the human body rather than detection of the whole body. Detection of Keypoints have potential applications in the pose estimation and consequently behavior analysis. Their architecture consist of two convolutional neural network where the first network extract features and gives the location of main joints of the human body in the form of heat map. While the second network is responsible for the associating the corresponding body joints. For the feature extraction, they used the classical VGG architecture. The frameworks like OpenPose are very helpful in skeleton extraction of the human body and potentially, it could be used in tracking framework. For example, the Bayesian framework proposed in \cite{ullah2016hog} works as a Keypoint tracker. Where any Keypoints like the position of head, or neck or any other body organ can be used to do tracking for longer time. Such Keypoints can be obtained from a variety of human pose estimation algorithms. For example, Sun et al. \cite{sun2019deep} proposed a parallel multi-resolution subnetworks for the human pose estimation. The idea of parallel network helps to preserver high resolution and yield high quality features maps that results better spatial Keypoints locations. Essentially, in such a setting, the detection module is replaced by \cite{cao2018openpose, sun2019deep}. In this regards, a global optimization approach like \cite{ullah2018directed} could be helpful for the accurate tracking in offline setting. By focusing only on pose estimation of humans, Fang et al. \cite{fang2017rmpe} proposed a top down approach where first the humans are detected in the form of bounding boxes and later, the joints and Keypoints are extracted through a regional multi-person pose estimation framework. Such a framework is helpful in not only in the localization and tracking of tracking in the scene, but also getting the pose information of all the targets in sequential fashion. For a robust appearance model that could differentiate between different targets, a sparse coded deep features framework was proposed in \cite{ullah2017hierarchical} that accelerate the extraction of deep features from different layers of a pre-trained convolution neural network. The framework is helpful in handling the bottleneck phenomenon of appearance modeling in the tracking pipeline. Alexander et al.\cite{alex2019pretraining} used transfer learning and fine-tuned ResNet to detect 22 joints of horse for the pose estimation. They used the data collected from 30 horses for the within domain and out of domain testing. The work by Mathis et al. \cite{mathis2018markerless} analyzed the behavior of mice during the experimental tasks of tail-tracking, and reach \& pull of joystick tasks. They also analyze the behavior of drosophila while it lays eggs. The classical way of inferring behavior is to perform segmentation and tracking \cite{ullah2018deep} first, and based on the temporal evolution of the trajectories, perform behavior analysis. However, approaches like \cite{nasirahmadi2017implementation} can be use to directly infer predefined actions and behaviors from the visual data. In addition to the visual data, physiological signals \cite{kanwal2019image, atlan2019frequency}, and acoustic signals \cite{cordeiro2018use} can be used to identify different emotional states and behavioral traits in farm animals. 

Compared to the existing methods, our proposed framework is focused on the extraction of the key joints of the Durac in an indoor setting. The visual data is obtained from a head mounted Microsoft Kinect sensors. Our proposed framework is inspired by \cite{psota2019multi} where a fully-convolutional stacked hourglass-shaped network is proposed that converts the image into a 16-channel space representing detection maps. For the part detection, the thresholds are set from 0.10 to 0.90. These threshold are used while evaluating the recall, precision, and F-measure metrics for both the vector matching and euclidean matching results. Such an analysis provides a detailed overview of the trade-offs between precision and recall while maintaining an optimal detection threshold. The loss function, the optimizer and the training details are also given in \ref{sec:model_arch}. The qualitative results are mentioned in \ref{sec:experiments} and the remarks are given in \ref{sec:conclusion} which concludes the paper.

%\cite{presti20163d}
The rest of the paper is organized in the following order. In section \ref{sec:proposed_approach} the proposed method is briefly explained including the Keypoints used in the experiment, the data filtration and annotation, and the augmentation. Model architecture is elaborated in section \ref{sec:model_arch}. The loss function, the optimizer and the training details are also given in \ref{sec:model_arch}. The qualitative results are mentioned in \ref{sec:experiments} and the remarks are given in \ref{sec:conclusion} which concludes the paper.

\begin{figure*}[h]
	\includegraphics[width=\textwidth]{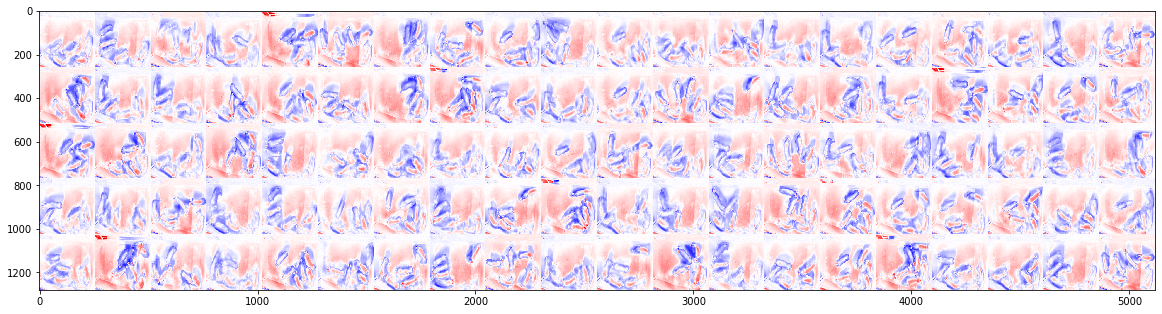}
	\caption{Image Data after K-Means Sampler}
	\label{fig:kmeans}
\end{figure*}
%\subsection{Key Points}

\section{Proposed Approach}
\label{sec:proposed_approach}

The block diagram of the network is given in Fig. \ref{fig:lossfunction}. It mainly consist of two encoder-decoder stacked back to back. The convolution neural network used in each encoder-decoder network is based on dense net. The network takes the input as the visual frame. In order to train the model, we annotate the data by first converting videos into individual frames, and then annotating each frame separately by specifying the important key points on the animal's body. After sufficient training, the model returns a 9x3 matrix for each frame, where each row corresponds to one keeping, the first two columns specify the x and y coordinates of the detected point, and the third column contains the confidence score of the model. After obtaining the x and y coordinates for each frame, we visualize these key points on each frame and stitch all the individual frames into a single video. A total of nine key points is being focused for each Duroc. Key points are as follows: Nose, Head, Neck, Right Foreleg, Left Foreleg, Right Hind leg, Left Hind Leg, Tail base, and tail tip. Key points and their connection is written in the CSV file for the annotator to read. Exact key points and their relations are shown in Table \ref{tb:keypoints}.
\begin{table}%[H]
	\centering
	\caption{Key Points Names}
	\label{tb:keypoints}
	\begin{tabular}{|c|c|c|}
		\hline
		\textbf{name} & \textbf{parent} & \textbf{swap} \\ \hline
		snout         &                 &               \\ \hline
		head          & snout           &               \\ \hline
		neck          & head            &               \\ \hline
		forelegL1     & neck            & forelegR1     \\ \hline
		forelegR1     & neck            & forelegL1     \\ \hline
		hindlegL1     & tailbase        & hindlegR1     \\ \hline
		hindlegR1     & tailbase        & hindlegL1     \\ \hline
		tailbase      &                 &               \\ \hline
		tailtip       & tailbase        &               \\ \hline
	\end{tabular}
\end{table}

In Fig. \ref{fig:lossfunction}, input images $x$ (top-left) are augmented (bottom-left) with various spatial transformations such as rotation, translation, scale, etc. In addition to that, they are followed by noise transformations such dropout, additive noise, blurring, contrast, etc. These transformations are applied to improve the robustness and generalization of the model. The ground truth explanations are then changed with coordinating spatial growths and used to draw the certainty maps $y$ for the key points and various hierarchical posture graph (top-right). The images $x$ are then passed through the network to produce a multidimensional array $g(f(x))$ a stack of images corresponding to the key point and posture graph confidence maps for the ground truth $y$. Mean squared error between the outputs for both networks $g(f(x))$ and $f'(x)$ and the ground truth data $y$ is then minimized (bottom-right), where $f'(x)$ indicates a subset of the output from $f(x)$. only those feature maps being optimized to reproduce the confidence maps for the purpose of intermediate supervision. The loss function is minimized until the validation loss stops improving—indicating that the model has converged or is starting to over fit to the training data \cite{graving2019deepposekit}. 

In the subsequent subsections, the data filtration, data augmentation and data annotation techniques used to produce the training data is explained. 

%This project aims to solve the problem of behavioral analysis of farm animals through movement tracking of skeletal key points on animal bodies. In order to solve this problem, 

%we train a Deep Neural Network which learns the distribution of Keypoints across an RGB frame. 

\subsection{Data Filtration}
The given data is in RGB format had different categories as per the number of Pigs. For example, there is a separate dataset for three pigs, six pigs, and 10 pigs. Each dataset has 2880 images. To get better and more accurate results, a larger dataset was required, hence more images need to be annotated. To overcome this problem, a video is created from the image data. After that, the frames were generated from the video. In order to save time and effort, while keeping in mind the original goal of accuracy, K-Means filter is applied to the frames, by which only 280 images are extracted from the larger dataset. The count is approximately 10 percent of the size of the original dataset.  K-Means sample configuration is as follows:
\begin{itemize}
	\item Batch size = 100
	\item clusters size = 100
	\item Reassignment Ratio = 0.01
	\item Tol = 0.0 
	\item verbose = True
\end{itemize}

Skeleton from the CSV file and Image data extracted after applying the K-Means sampler, combined data is stored in .h5 format. Complete dataset after diving into 100 unique clusters is shown in Fig. \ref{fig:kmeans}.

\subsection{Data Annotation}
\begin{figure}
	\centering
	\includegraphics[scale=0.40]{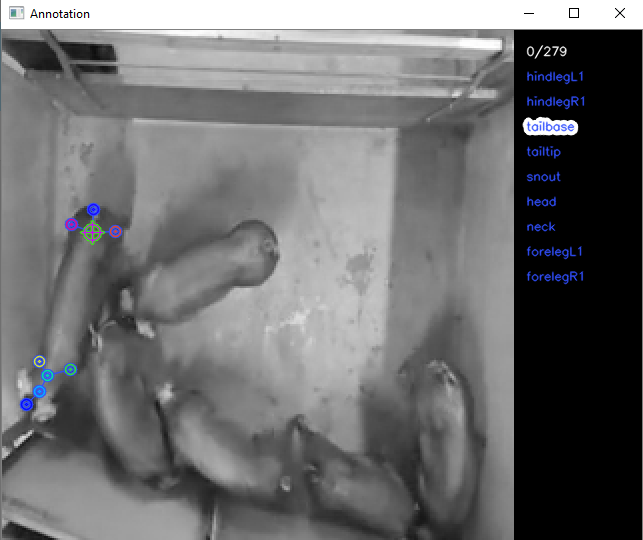}
	\caption{Data Annotator}
	\label{fig:da}
\end{figure}

Data annotator developed by Jake Graving is used to annotate the dataset. It provides a simple graphical user interface that reads key points data from the CSV file and saves the data in .h5 format once the annotation is completed. The data annotator is show in Figure \ref{fig:da}.
DeepPoseKit works with augmenters from the imgaug package. We are using spatial augmentations 
with axis flipping and affine transforms. deepposekit.augment.FlipAxis takes the DataGenerator as an argument to get the key point swapping information defined in the annotation set. When the images are mirrored key points for left and right sides are swapped to avoid "confusing" the model during training.
\iffalse
\subsection{Augmentation pipeline}
DeepPoseKit works with augmenters from the imgaug package. We are using spatial augmentations 
with axis flipping and affine transforms. deepposekit.augment.FlipAxis takes the DataGenerator as an argument to get the key point swapping information defined in the annotation set. When the images are mirrored key points for left and right sides are swapped to avoid "confusing" the model during training. After loading an image-key-points pair, and applying augmentation, some random augmentations are show in Figure \ref{fig:AP}.
\begin{figure}
	\centering
	\includegraphics[scale=0.6]{Augmentation.png}
	\caption{Augmentation pipeline}
	\label{fig:AP}
\end{figure}
\fi

\section{MODEL ARCHITECTURE}
\label{sec:model_arch}

The proposed framework is based on Stacked densenet, an efficient multi-scale deep-learning model. A quick GPU-based pinnacle identification calculation for evaluating Keypoint areas with sub-pixel exactness. These advances improve preparing speed greater than 2x with no accuracy in exactness contrasted with at present accessible strategies~\cite{graving2019deepposekit}. Densenet is a densely Connected Convolutional Networks~\cite{huang2017densely}. DenseNet can be seen as the next generation of convolutional neural network that are capable of increasing the depth of model with every decreasing the number of parameter. 

\iffalse
The graphical depiction of densenet architecture is shown in Fig. \ref{fig:model_architecture}. 
\begin{figure}
	\centering
	\includegraphics[scale=0.5]{Architecture.png}
	\caption{DenseNet with 5 layers with an expansion of 4~\cite{huang2017densely}}
	\label{fig:model_architecture}
\end{figure}
\fi

\subsection{Loss function \& Optimizer}
We have used the callback function using \emph{ReduceLROnPlateau}. ReduceLROnPlateau automatically reduces the learning rate of the optimizer when the validation loss stops improving. This helps the model to reach a better optimum at the end of training. Following are the parameters for this function:
\begin{itemize}
	\item Monitor: \textit{val\_loss}
	\item Factor: 0.2
	\item Verbose: 1
	\item Patience: 20
\end{itemize}

Mathematically, the loss function is:
\begin{equation}
	L(x, y) = \frac{1}{n} \sum_{i}^{n} ((g(f(x))_i-y_i)^2 + (f'(x)-y_i)^2) 
\end{equation}
%An illustration of the model training process for our Stacked DenseNet model our project is show in Figure \ref{fig:lossfunction}. 

\subsection{Training}
\subsubsection{Training Model}
\begin{figure}[h]
	\centering
	\includegraphics[width=8cm,height=7cm]{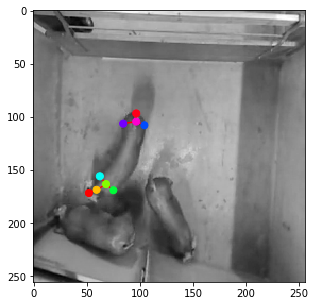}
	\caption{Data Generator}
	\label{fig:data_generator1}
\end{figure}

While training a model on three pigs’ data, first a test was run for data generators. Figure \ref{fig:data_generator1} is the image of the very first image from the dataset. 

\subsubsection{Training Generator}

\begin{figure}[h]
	\centering
	\includegraphics[scale=0.35]{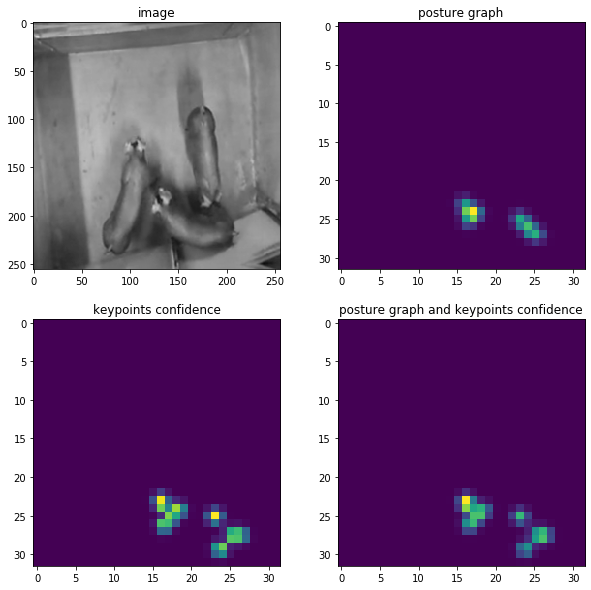}
	\caption{Training Generator}
	\label{fig:data_generator}
\end{figure}

Creating a TrainingGenerator from the DataGenerator for training the model with annotated data is an important factor. The TrainingGenerator uses the DataGenerator to load image-keypoints pairs and then applies the augmentation and draws the confidence maps for training the model. Figure \ref{fig:data_generator} shows the training generator for one randomly picked frame.

The \textbf{validationSplit} argument defines how many training examples to use for validation during training. If a dataset is small (such as initial annotations for active learning), we can set this to \textbf{validationSplit=0}, which will just use the training set for model fitting. However, when using callbacks, we made sure to set monitor="loss" instead of monitor="valloss". To make sure the Training Generator is working, the following are some output visuals.

\subsubsection{Reduce Learning Rate}
Reduce learning rate parameters saves useless resource utilization and model overfeeding. For this particular reason, the parameter is set to reduce the learning rate by 0.2 if the loss does not improve after 20 iterations. 

\subsubsection{Early Stopping}
Another parameter that is used to prevent resource exploitation is Early Stopping. Patience is set 100 iterations that are, training would stop automatically if the loss does not improve after 100 iterations. 

\subsubsection{Training Results}
Training started at a loss of 220, after running 400 iterations, the loss stopped showing improvement at 4.5. In the test case, when given the same video from which dataset was generated, very accurate results are produced. Below are some results of the iterations after running 50 iterations on the dataset.

\begin{figure}[h]
	\centering
	\includegraphics[width=7.5cm,height=6.5cm]{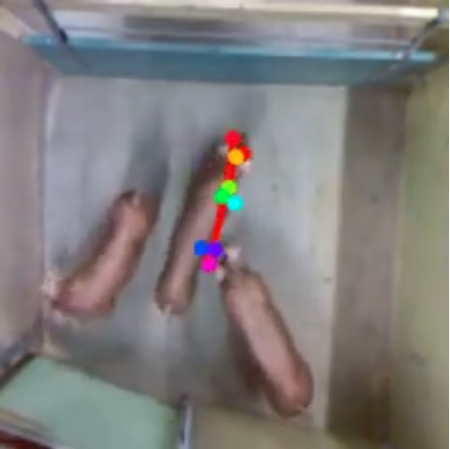}
	\caption{Results after 50 Iterations}
	\label{fig:training_results1}
\end{figure}

\begin{figure}[h]
	\centering
	\includegraphics[width=7.5cm,height=6.5cm]{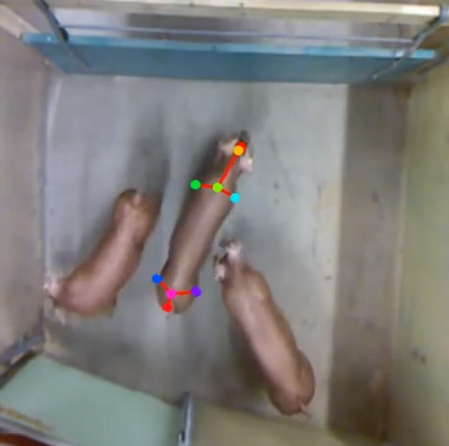}
	\caption{Results after 400 Iteration}
	\label{fig:g_results1}
\end{figure}

\begin{figure}[h]
	\centering
	\includegraphics[width=7.5cm,height=6.5cm]{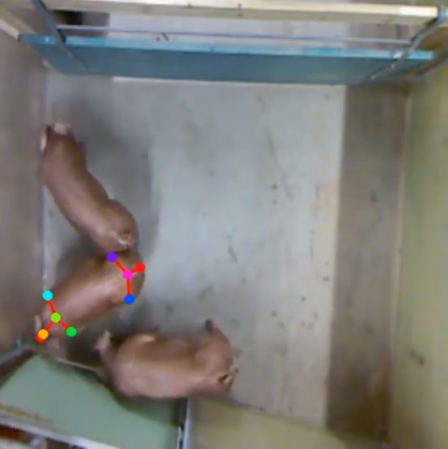}
	\caption{Results after 50 Iterations}
	\label{fig:g_results2}
\end{figure}

In Fig. \ref{fig:training_results1}, since the model is not properly trained, model is detecting very rough key points. Hardly the key points on head section is detected. On the tail section key points are not detected at all. Since the model showed a great tendency to learn, hence after 400 iterations, the results are shown in Fig. \ref{fig:g_results1} and Fig. \ref{fig:g_results2}. Clearly difference between the accuracy can be seen between Fig. \ref{fig:training_results1} and Fig. \ref{fig:g_results1} as the result is generated on exactly the same frame.

Result was generated on total of 2880 frames, in Figure \ref{fig:g_results2} is showing pigs with different locations, yet very perfect key points detection.

\subsection{Outlier Frame}
\begin{figure*}
	\includegraphics[width=\textwidth]{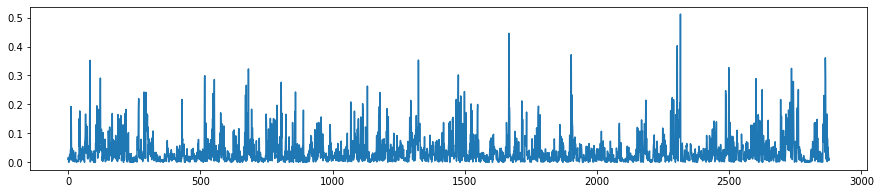}
	\caption{Confidence Difference}
	\label{fig:confidence_difference}
\end{figure*}

\begin{figure*}
	\includegraphics[width=\textwidth]{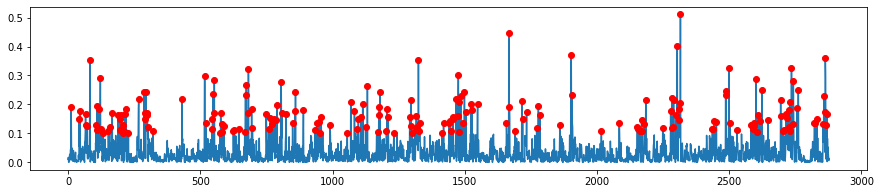}
	\caption{Outliers}
	\label{fig:outliers}
\end{figure*}

Using the confidence scores and the temporal derivatives, we detected the potential outliers and added them to the annotation set. The confidence difference is show in Fig. \ref{fig:confidence_difference}. Using the \textit{scipy.signal.find\_peaks} package, we detected the outliers and the generated output is show in Fig. \ref{fig:outliers}.

\iffalse
\subsection{Object Tracking}
During the prediction stage, the data is stored in the form of a numpy array in a separate file. The data is stored in the form of a 9x3 matrix for each frame. In this case, because of the total of 2880 frames, the shape of a numpy array is (2880, 9, 3). Each instance of the matrix stores the x and y-axis of a certain key point as well as the confidence of the prediction.

In order to do object tracking using Kalman Filter, it was important to find out the central key point of the pig. To do so, we calculated a central point using the existing four key points which are two forelegs and hindlegs. The center between two points $FLR(x1, y1)$ and $HLL(x2, y2)$ is given by:
\begin{equation}
M = ((xA+xB)/2 ,  (yA+yB)/2)
\end{equation}
The point M further used to predict the movement of the pig. By giving the starting point to the measurement matrix and providing another two positions as an observation transition matrix, the following results are achieved. The Red line indicates the predicted place whereas the yellow point indicates the actual movement of the object. 

\fi

\section{Experiments}
\label{sec:experiments}

The proposed framework is implemented in Python with the support of Keras backend by Tensorflow. The processing is performed on Nvidia P-100 with 32 GB RAM.

\section{Conclusion}
\label{sec:conclusion}
Since we had a constraint environment in the data set, we had almost similar backgrounds, camera angels, and color of objects (pigs), hence the model showed larger tendency to learn as compare to training the model on the dataset gathered for the objects in the wild. By applying the K-Means sampler, we were able to train the model on a bigger data distribution, hence eliminating redundant data frames. This helped us to filter out unique data frames. By using StackedDensenet pre-trained weights to train our model, even providing the training set of just 280 images, we got very promising results.

\section{APPENDICIES}
\label{sec:appendicies}

In this section, we mentioned some resources that we used and the techniques we implied that did not work out as expected. Following the summary of the process and the work-done that resulted in a deadlock. 

\subsection{Environment}
The first step we did was creating a Conda environment on Ubuntu operating system. The second step was to install all the dependencies and packages that were needed in the environment for running the pose estimation tensor flow. The code we used was from this Github repository \cite{tf-pose}. We were able to do pose estimation on human images and human videos using the CMU model by running this very code-base. 

\subsection{Keypoints}
A total of 5 key points were being tracked for the pigs. Key points that were being tracked were Nose, Neck, Tail, Left Ear, Right Ear.  We had data of 2800 images of pigs. So the next step was to annotate the data.

\subsection{Data Annotations}
We used an online annotator for annotating the dataset. The annotator can be found at this link \cite{ImgLab}. We had to upload one image at a time in the annotator. We had to select all the key points and set the bounding box on every pig in the image. The annotator then saved all the labels in the COCO JSON format.

\subsection{Training Model}
We started to train the model on the data we gathered from the annotator. We trained the model after doing necessary customization in the code-base available at the Github repository \cite{Baseline}. At the start of the training, the loss was at 270 and by the end of 140th iteration, the loss was at 22. Hence the model showed a great tendency of learning. After the training, the results were in .ckpt format which we could not use in the existing codebase. Although we were able to generate Protobuf format due to the difference in the model architecture, we could not generate any results while using it.

%-------------------------------------------------------------------------------%------------------------------------------------------------------------------------------------------------------%

%\section*{Acknowledgments}
%This research was funded by Norwegian Ministry of education and research and Hainan provincial scientific collaboration project under grant KJHZ2015-23.
%------------------------------------------------------------------------------------------------------------------%

{\small
\bibliographystyle{IEEEbib}
\bibliography{CleanNew}
%\bibliography{strings,refs,external}
}
\iffalse
\begin{figure}[!hb]
	\includegraphics[width=0.2\columnwidth]{logo.png}
	\caption{IS\&T logo.}
	\label{Figure:logo}
\end{figure}

\fi
\end{document}